%% file: main.tex
\documentclass[10pt,twocolumn,letterpaper]{article}

\usepackage{cvpr}              
\input{preamble}
\definecolor{cvprblue}{rgb}{0.21,0.49,0.74}
\usepackage[pagebackref,breaklinks,colorlinks,allcolors=cvprblue]{hyperref}

\def\paperID{45} 
\def\confName{CVPR}
\def\confYear{2026}

\title{SAM for Robust Mitochondria Instance Segmentation in Fluorescence Microscopy}

\author{Suyog Jadhav, Dilip K. Prasad, Krishna Agarwal\\
{\small
UiT The Arctic University of Norway
} \\
{\scriptsize \texttt{
\{suyog.s.jadhav, dilip.prasad, krishna.agarwal\}@uit.no
}}
}

\begin{document}
\maketitle
\input{sec/0_abstract}    
\input{sec/1_intro}
\input{sec/2_related_works}
\input{sec/3_finalcopy}
{
    \small
    \bibliographystyle{ieeenat_fullname}
    \bibliography{main}
}


\end{document}

%% file: sec/0_abstract.tex
\begin{abstract}
The morphological analysis of mitochondria in fluorescence microscopy (FM) is crucial for understanding cellular health, energy production, and metabolic regulation. While foundation models like the Segment Anything Model (SAM) have revolutionized natural image segmentation, their direct application to FM is hindered by a significant domain shift characterized by diffraction-limited resolution, low contrast, and complex overlapping organelle networks. Furthermore, the development of robust models is bottlenecked by a severe lack of high-quality, manually annotated instance segmentation datasets for mitochondria. In this paper, we propose a scalable solution to this data scarcity by finetuning SAM exclusively on synthetically generated FM data. We simulate realistic mitochondria data and emulate the optical properties of fluorescence microscopes to create a large-scale annotated dataset. We evaluate our fine-tuned model on a curated dataset of real, manually annotated FM images. Qualitative and quantitative analyses demonstrate that our synthetically fine-tuned model improves precision and average dice score over strong baselines. This work establishes the potential of simulation-assisted training for FM instance segmentation.
\end{abstract}

%% file: sec/1_intro.tex
\section{Introduction}
\label{sec:intro}

Mitochondria are highly dynamic organelles whose morphology, distribution, and function are tightly linked to cellular metabolism, aging, and various pathologies such as neurodegenerative diseases. Fluorescence microscopy (FM) remains the gold standard for visualizing these structures in live-cell environments due to its non-destructive nature, which allows researchers to capture temporal dynamics under various biological stimuli.

Automated, highly accurate instance segmentation of these organelles is highly desirable to enable fine-grained downstream analyses, such as temporal tracking of mitochondrial migration, fusion, and fission. However, FM suffers from inherent image quality challenges, including variable noise, low Signal-to-Background Ratios (SBR), and diffraction blurring. Further, mitochondria frequently form dense, interconnected clusters, and the inherently low-contrast nature of FM often leads traditional thresholding and contour-based methods to erroneously interpret a cluster. This leads to fragmentation in some cases, where each branch of a junction gets treated as a new mitochondrion. In other cases, it leads to a large cluster being interpreted as a single continuous structure.

This problem is further exacerbated by the severe lack of high-quality, annotated instance segmentation datasets in the public domain. While vast quantities of raw FM data exist, manual annotation of overlapping, low-contrast mitochondrial networks is an arduous, and often ambiguous process \cite{physeg}. Due to these reasons, training fully supervised deep learning models for instance segmentation in this specific domain has been challenging.

The introduction of vision foundation models, most notably the Segment Anything Model (SAM) \cite{kirillov2023segment}, has led to significant improvements across various image analysis tasks. While adaptations like $\mu$SAM (micro-SAM) \cite{archit2025segment} have attempted to bridge the gap into the microscopy domain, their reliance on existing datasets limits their application on data-deficient domains such as fluorescence microscopy. Other specialized tools like Nellie \cite{lefebvre2025nellie} perform better in comparison, but struggle with false positives and fragmentation in low-contrast regions, which can lead to less reliable quantitative and temporal analysis downstream. Further, the pixel-classification approach used by these methods does not accurately model mitochondria in dense networks. The high amount of overlaps in such networks necessitates an approach that models these overlaps correctly.

We propose a new instance segmentation model better suited for fluorescence microscopy images than the existing state-of-the-art approaches. We showcase the achieved improvements through qualitative and quantitative comparisons on manually annotated data. We further demonstrate the applicability of our approach in analysing biological experiments using a case study on publicly available mitophagy data.

Our primary contributions are:
\begin{itemize}
    \item A simulation-supervised training paradigm to address the problem of ground truth deficiency
    \item Finetuned SAM model that improves mean dice score and precision compared to existing state-of-the-art methods
    \item Improved instance delineation in dense networks with reduced fragmentation and better overlap handling
\end{itemize}

%% file: sec/2_related_works.tex
\section{Related Work}
\label{sec:related-work}

\textbf{$\mu$SAM:} The introduction of SAM has inspired numerous adaptations for medical and biological imaging. CellSAM \cite{cellsam} and Cellpose-SAM \cite{cellposesam} have adapted the architecture for cellular segmentation, while $\mu$SAM extends SAM to various light and electron microscopy modalities. $\mu$SAM also introduces an Automatic Instance Segmentation (AIS) mode using an additional decoder that predicts object centers to seed a watershed algorithm. This approach is effective and works for well-separated instances (for e.g., light microscopy images of a tissue culture showing individual cells). But since it limits each pixel to a single instance, it leads to over-simplification in dense mitochondrial networks when applied to fluorescence microscopy. 

\textbf{Nellie:} Nellie \cite{lefebvre2025nellie} is a recently introduced tool for automated organelle segmentation and tracking. It utilises regional similarities to classify pixels into individual instances using decision trees. Further, it uses Frangi vesselness filtering \cite{frangi1998multiscale} to enhance the foreground details and reduce the influence of background regions. While Nellie is specialized for fluorescence microscopy data, we found it to be prone to producing a high number of false positives in low-contrast regions of the image, leading to fragmentation. These false positives can lead to inconsistency issues in downstream tasks such as automated analysis and tracking.

\textbf{Synthetic Data for Bioimage Analysis:} Generating synthetic microscopy images has been explored for semantic segmentation. Tools like PhySeg \cite{physeg} utilize physics-based rendering to simulate subcellular structures such as mitochondria, vesicles, actins etc. We adapt and expand upon this work to further generate instance-level ground truths. This enables the finetuning of instance segmentation models without having to rely on manual annotation.

\textbf{Modelling Overlaps:}
The existing state of the art methods in this domain, $\mu$SAM and Nellie, rely on a per-pixel classification approach. In a per-pixel classification approach, individual pixels are grouped into multiple clusters using some type of similarity metric. This approach limits each pixel to belonging to no more than one instance. In a typical fluorescence microscopy image, mitochondria are seen forming large networks with many overlaps and weak delineation between the individual instances. This makes the task of instance segmentation challenging. Modeling such overlaps with a per-pixel classification approach is inherently limiting. Promptable segmentation models such as SAM allow for a better approach, where each prompt is processed independently of the others. This makes it possible to model overlaps more naturally, since the combined output from all the prompts allows more than one instance to overlap and share one or more pixels. The use of such promptable approach has seen limited use in the domain.

\section{Methodology}

We employ a procedural simulation pipeline to generate synthetic image-mask pairs. Each image contains multiple mitochondria, each of which has a ground truth segmentation mask associated with it. We use this dataset for finetuning SAM. For inference, we use a modified automatic mask generation approach, that focuses on gathering prompts only from the foreground region.

\subsection{Simulation Pipeline}

\begin{figure*}[htbp]
\centering
\begin{subfigure}[b]{0.3\textwidth}
  \centering
  \includegraphics[width=.8\linewidth]{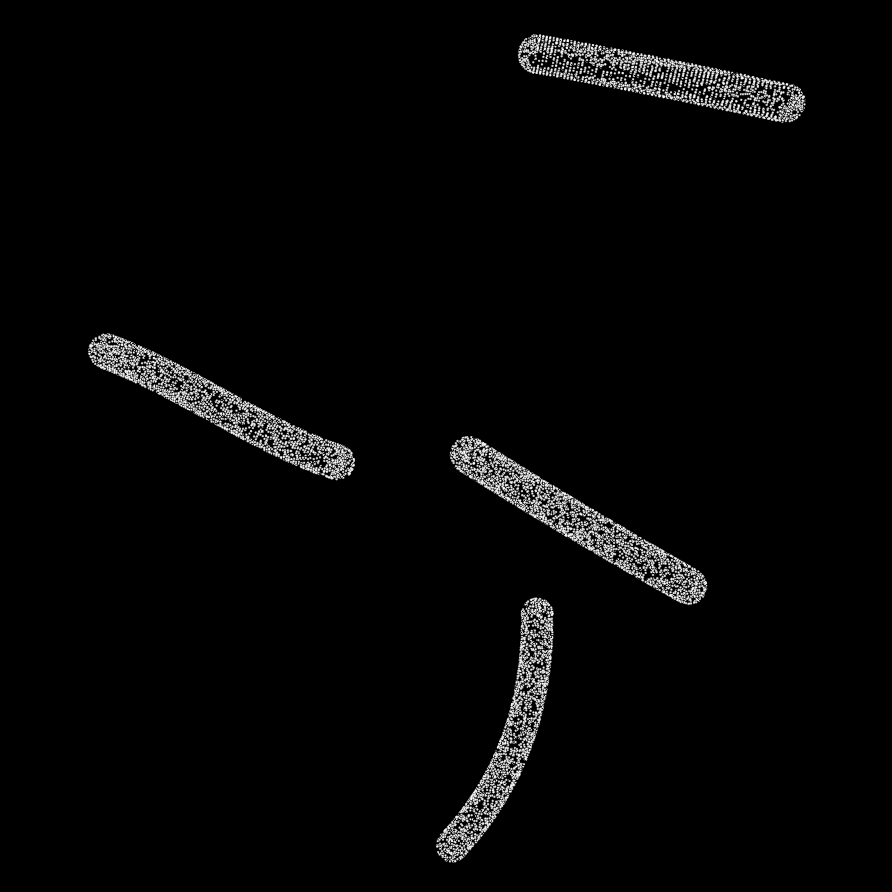}  
  \caption{Simulated point cloud}
\end{subfigure}
\begin{subfigure}[b]{0.3\textwidth}
  \centering
  \includegraphics[width=.8\linewidth]{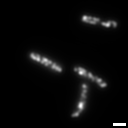}
  \caption{Convolution with PSF}
\end{subfigure}
\begin{subfigure}[b]{0.3\textwidth}
  \centering
  \includegraphics[width=.8\linewidth]{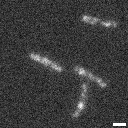}  
  \caption{Poisson noise addition (SBR=3)}
\end{subfigure}
\caption{Data simulation pipeline used for generating the synthetic data. Scale bar is 1 $\mu m$.}
    \label{fig:mito_sim}
\end{figure*}

We model individual mitochondria as thick cylindrical shapes fitted onto an $n$-point Bezier curve, with hemispherical caps. To create a sample, we generate multiple simulated mitochondria (ranging from 4 to 16 per $128 \times 128$ pixel crop) placed randomly in a 3D coordinate space. 

To convert this 3D point cloud into a realistic fluorescence microscopy image, we convolve the points with a Point Spread Function (PSF). For this, we utilize an optimized implementation of the Gibson-Lanni PSF model \cite{gibson1992experimental} used by \cite{physeg}. The simulation models a numerical aperture of 1.42, a magnification of $100\times$, and an emission wavelength of 660 nm.

\begin{table}[htbp]
\centering
\caption{Parameters used for simulating fluorescence microscopy images of mitochondria.}
\label{tab:sim_params}
\begin{tabular}{ll}
\toprule
\textbf{Parameter} & \textbf{Value} \\
\midrule
Spatial Range ($x, y$) & $[-5000, 5000]$ nm \\
Axial Range ($z$) & $[0, 500]$ nm \\
Mitochondria/Sample & 4 to 16 \\
Mitochondria Length & $[500, 2500]$ nm \\
Mitochondria Width & $[200, 500]$ nm \\
SBR Values Range & $[2, 6]$ \\
Pixel size & 80 nm \\
\bottomrule
\end{tabular}
\end{table}

To mimic the effect of multiple focus depths, we take 10 slices through the simulated $z$-space and then perform a maximum intensity projection to obtain a 2D image. This introduces realistic out-of-focus blur in the produced images. Finally, to mimic the realistic noise distribution of fluorescence microscopy images, we add Poisson noise. The noise level is determined by randomly sampling a Signal-to-Background Ratio (SBR) between 2 and 6. For larger fields of view, the base $128 \times 128$ crops are tiled (for e.g., tiling 4 times yields a resolution of $512 \times 512$) to introduce spatial variance and edge complexities. A summary of the simulation parameters is provided in Table \ref{tab:sim_params}. These parameters are used to generate a training dataset consisting of a total of 10000 images of resolution 512x512.

\subsection{Model Finetuning}

The standard SAM architecture consists of an image encoder (ViT \cite{vit}), a prompt encoder, and a lightweight mask decoder. During training, for each image, we sample a fixed number of points at random from the total foreground area to use as prompts to SAM. The outputs are then compared with the segmentation masks of the instances the point lies on. In case of a point lying on multiple overlapping instances, the output is compared with all of the overlapping segmentation masks, and the mask producing the least loss is used for backpropagation. The model is optimized using standard Dice loss, computed as shown in eq. \ref{eq:dice_loss}.

\begin{equation} \label{eq:dice_loss}
Loss_{dice}(X, Y) = 1-\frac{2 |X \cap Y|}{|X| + |Y|}    
\end{equation}

By exposing the model to thousands of synthetic overlaps and junctions, it learns to extract only the prominent structure corresponding to the prompt point, implicitly learning to ignore other occluding mitochondria in the network. Further, since each point prompt is processed separately from each other, an individual segmentation mask is produced per prompt. This allows each pixel to be shared between one or more instances. This method encodes overlaps more naturally compared to a per-pixel classification approach.

Finally, different models were trained with variation in their sizes and finetuning strategies. These models showed noticeably different performances. Table \ref{tab:ablation} shows the results of SAM trained on our synthetic dataset with/without finetuning the image encoder, and with ViT-base or ViT large as the backbone. The metrics are calculated on the manually annotated test dataset. For more details about the metrics and the test dataset used, refer to section \ref{sec:experiments}. From the table, we can see that model size correlates inversely with model performance, with smaller ViT-base model generally performing better than ViT-large. Finetuning both image encoder and the mask decoder also showed a decrease in model performance compared to finetuning just the mask decoder. The best model was thus observed to be ViT-base with its image encoder frozen during training and only the mask decoder. This is the model we use for evaluation in the following sections. 

\begin{table}[htbp]
\caption{Comparing model finetuning strategies (MD=Mask Decoder; IE=Image Encoder). Best results for each column are emphasized.}
\label{tab:ablation}
\resizebox{\columnwidth}{!}{
    \begin{tabular}{cccccc}
    \hline
    \textbf{\begin{tabular}[c]{@{}c@{}}SAM \\ Backbone\end{tabular}} & \textbf{\begin{tabular}[c]{@{}c@{}}Parts\\ Finetuned\end{tabular}} & \textbf{Precision} & \textbf{Recall} & \textbf{\begin{tabular}[c]{@{}c@{}}Dice\\ (NZ)\end{tabular}} & \textbf{\begin{tabular}[c]{@{}c@{}}Mean \\ Dice\end{tabular}} \\ \hline
    ViT-base                                                         & MD                                                                 & \textbf{0.555}     & \textbf{0.282}  & \textbf{0.555}                                               & \textbf{0.483}                                                \\
    ViT-base                                                         & MD, IE                                                             & 0.460              & 0.174           & 0.504                                                        & 0.434                                                         \\
    ViT-large                                                        & MD                                                                 & 0.530              & 0.199           & 0.512                                                        & 0.450                                                         \\
    ViT-large                                                        & MD, IE                                                             & 0.482     & 0.185           & 0.525                                               & 0.437                                                \\ \hline
    \end{tabular}
}
\end{table}

\subsection{Inference}
During inference, we employ an automated pipeline to generate instance segmentations. Firstly, since microscopy images can look significantly different based on the pixel size, a preprocessing step is employed to match the field of view of the input images to those used during training. Images are first up/downsampled to match the training pixel size of 80 nanometers, followed by taking 512x512 patches from the image. Outputs from individual patches are then combined to get the final output.
To obtain the output, we use a modified automatic mask generation approach. We first estimate an approximate foreground by performing a binary thresholding operation on the input image. A large number of points is then sampled from this foreground region to be used as prompts for the model. Following this initial mask prediction, a post-processing step is performed to refine the outputs. Specifically, we get rid of highly fragmented masks by performing connected component analysis and keeping only the largest component for each mask. Masks that are too small (less than 10 pixels in area) are also filtered out, as well as masks that have an overlap higher than 0.5 with any neighboring masks. This reduces redundant predictions in dense mitochondrial networks.

\section{Experiments} \label{sec:experiments}

\subsection{Experimental Setup}

\begin{figure*}
\centering
\begin{subfigure}[b]{0.45\textwidth}
  \centering
  \includegraphics[width=\linewidth]{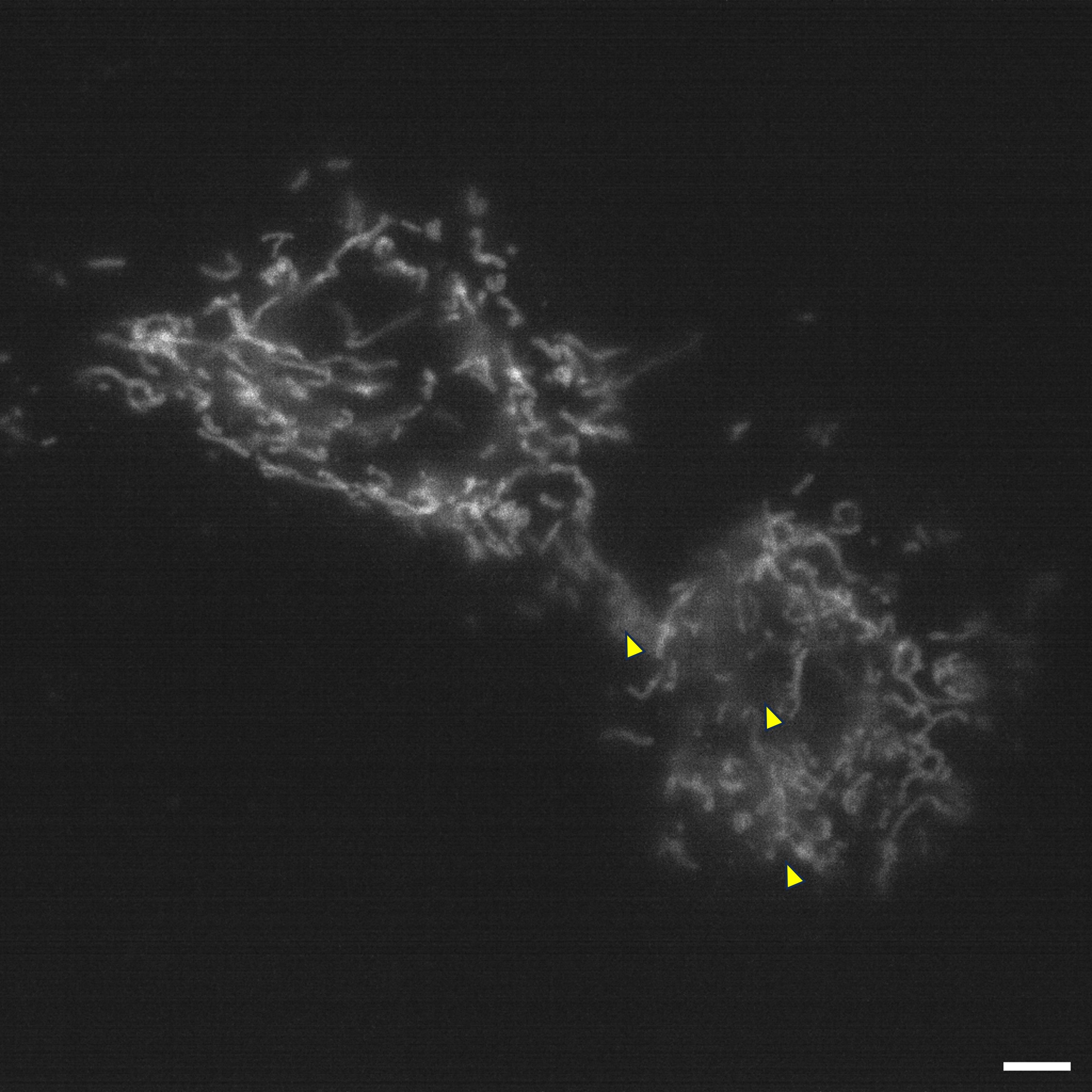}
  \caption{}
\end{subfigure}
\begin{subfigure}[b]{0.45\textwidth}
  \centering
  \includegraphics[width=\linewidth]{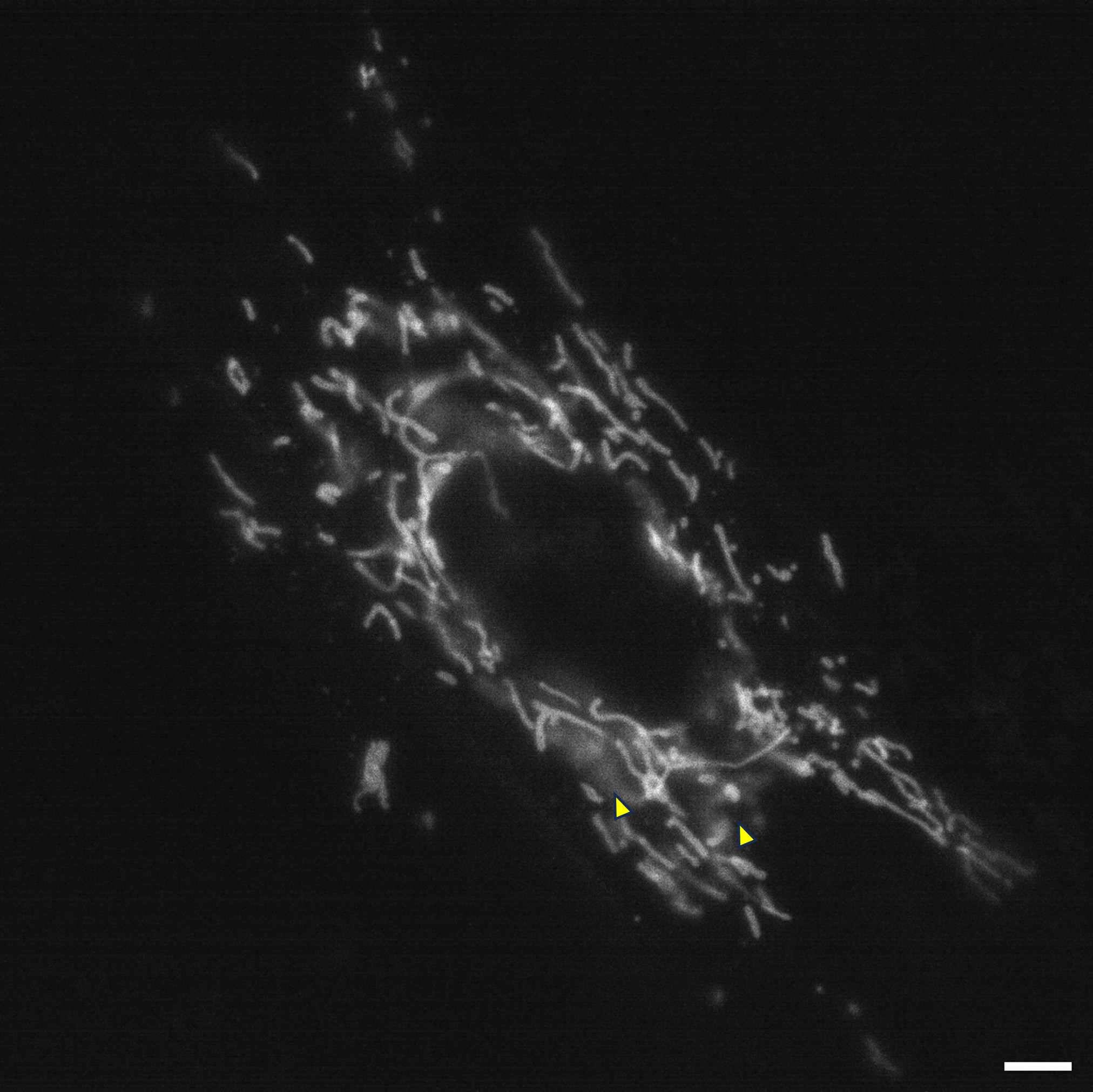}
  \caption{}
\end{subfigure}
\caption{(a) Test image 1, and (b) Test image 2. Scale bar is 5 \textmu m. Both images are sourced from the evaluation data publicly made available in \cite{physeg}. Yellow markers denote challenging low-contrast regions in the image.}
    \label{fig:test_images}
\end{figure*}

We compare our synthetically fine-tuned model against two of the current state-of-the-art baselines:
\begin{enumerate}
    \item \textbf{Nellie:} An automated organelle segmentation tool specifically designed for 2D/3D live-cell microscopy. It is trained on FM images. We use the publicly available plugin provided by the authors for our evaluation.
    \item \textbf{$\mu$SAM:} A SAM variant fine-tuned on multiple large-scale biological datasets. $\mu$SAM is not originally trained on FM images. We fine-tune $\mu$SAM on our synthetic dataset using the publicly available finetuning code provided by the authors. Being a method also based on SAM, $\mu$SAM serves as a good comparison of the training methodology itself. 
\end{enumerate}

Due to the lack of publicly available annotated datasets, we manually annotate two fluorescence microscopy images from the publicly available dataset from PhySeg \cite {physeg} to use as our test dataset. The two test images, shown in figure \ref{fig:test_images} have slightly different noise characteristics, with test image 1 being noisier than test image 2. Challenging low-contrast regions in both images are highlighted with yellow markers. Ground truth annotations were manually created using annotation tools in Napari \cite{napari}.

We evaluate the models based on Precision, Recall, and Dice score. The ground truth masks are matched with the predictions by calculating the overlap between each of the masks against each of the ground truth masks, and choosing the best match for each ground truth mask. Further, it is ensured that each "match" has a dice score of $\ge 0.5$  against their corresponding ground-truth instance. All the predictions that have a valid match with a ground truth instance are considered True Positives (TP). Any predictions that are not matched with a ground truth instance are considered False Positives (FP). Finally, any ground truth instances that are not matched successfully are considered False Negatives (FN). Precision and Recall are then calculated as shown in equations \ref{eq:prec} and \ref{eq:reca}.

\begin{equation}
    Precision = \frac{TP}{TP + FP}
    \label{eq:prec}
\end{equation}

\begin{equation}
    Recall = \frac{TP}{TP + FN}
    \label{eq:reca}
\end{equation}

Any unmatched instances are assigned a dice score of zero. We report the standard Mean Dice across all predictions (Mean Dice) as well as the Non-Zero Dice (Dice NZ), which is computed by taking into account the average dice score only for the instances with a dice score of $>0$. The dice score between any two segmentation masks $X$ and $Y$ is calculated as shown in eq. \ref{eq:dice_score}.

\begin{equation} \label{eq:dice_score}
Dice(X, Y) = \frac{2 |X \cap Y|}{|X| + |Y|}    
\end{equation}

Additionally, to limit the influence of background noise on the segmentation quality, any predicted segmentation masks with an area of smaller than 10 pixels is rejected before evaluation. While mitochondria lengths and diameter vary across cell types, they are commonly between 1-2 $\mu m$ long, and 0.5-1 $\mu m$ in diameter \cite{hom2009morphological}, although the diameter has been observed to be as low as 0.1 $\mu m$ in some rare cases \cite{jones1975ultrastructure}. From a top-down view, this makes the smallest mitochondrion roughly around 0.5 $\mu m^2$, or 0.1 $\mu m^2$ (in the exceptional case) in area. An area threshold of 10 pixels is physically equivalent to 0.064 $\mu m^2$ in the test images (owing to the pixel size of 80 nanometers), which is lower than the smallest expected mitochondrion, and hence not likely to filter out any actual mitochondria.

\subsection{Quantitative Results} \label{sec:quant}

Tables \ref{tab:quant_results1} and \ref{tab:quant_results2} showcase the metric scores calculated on the outputs produced by each of the methods.

\begin{table}[htbp]
\centering
\caption{Quantitative results on test image 1. Best results for each column are emphasized.}
\label{tab:quant_results1}  
\resizebox{\columnwidth}{!}{
    \begin{tabular}{lcccc}
    \toprule
    \textbf{Method} & \textbf{Dice (NZ)} & \textbf{Mean Dice} & \textbf{Precision} & \textbf{Recall} \\
    \midrule
    Nellie & 0.405& 0.093& 0.070& \textbf{0.294}\\
    
    $\mu$SAM & \textbf{0.512}& 0.334& 0.329& 0.257\\
    \textbf{Proposed} & 0.510& \textbf{0.428}& \textbf{0.444}& 0.252\\  

    
    \bottomrule
    \end{tabular}%
}
\end{table}

\begin{table}[htbp]
\centering
\caption{Quantitative results on test image 2. Best results for each column are emphasized.}
\label{tab:quant_results2} 
\resizebox{\columnwidth}{!}{
    \begin{tabular}{lcccc}
    \toprule
    \textbf{Method} & \textbf{Dice (NZ)} & \textbf{Mean Dice} & \textbf{Precision} & \textbf{Recall} \\
    \midrule
    Nellie & 0.579& 0.159& 0.189& \textbf{0.678}\\
    
    $\mu$SAM & 0.537& 0.442& 0.517& 0.298\\
    \textbf{Proposed} & \textbf{0.600}& \textbf{0.538}& \textbf{0.667}& 0.312\\

    
    \bottomrule
    \end{tabular}%
}
\end{table}

Across both test images, Nellie consistently achieves the highest Recall (0.294 and 0.678) but suffers from drastically low Precision (0.070 and 0.189) and Mean Dice (0.093 and 0.159). 

Conversely, $\mu$SAM exhibits notably better Precision (0.334 and 0.517) and Mean Dice (0.329 and 0.442) than Nellie, but its Recall drops significantly(0.257 and 0.298). This could be explained by the seeded watershed approach employed by $\mu$SAM not being able to efficiently delineate between the boundaries of individual mitochondria in a dense network, leading to overlapping mitochondria being interpreted as a single continuous structure and thereby missing some of the constituent individual instances.

Our proposed model bridges this gap effectively. It achieves the highest  Mean Dice (0.428 and 0.538) across both test images, indicating a comparatively better structural alignment with the ground truth. The non-zero dice score is the best amongst the three methods for test image 2 (0.600), whereas for test image 1, $\mu$SAM narrowly overtakes our model (0.512 Vs 0.510). Further, the value of mean dice score is, as expected, lower than the non-zero dice score for all the metrics. The drop between the two values, however, is the least significant for our model, indicating a reduction in inaccurate detections (detections with a dice score of 0) compared to other methods. The proposed model also consistently achieves the highest Precision scores (0.444 and 0.667), which correlates with the reduction in inaccurate detections. Lastly, the recall scores (0.252 and 0.312) are comparable to those produced by $\mu$SAM (0.257 and 0.298), and are noticeably lower than Nellie (0.294 and 0.678).

\subsection{Qualitative Analysis}
\label{sec:qual}
\begin{figure*}[htbp]
  \centering

\begin{subfigure}[b]{\textwidth}
  \centering
  \includegraphics[width=\linewidth]{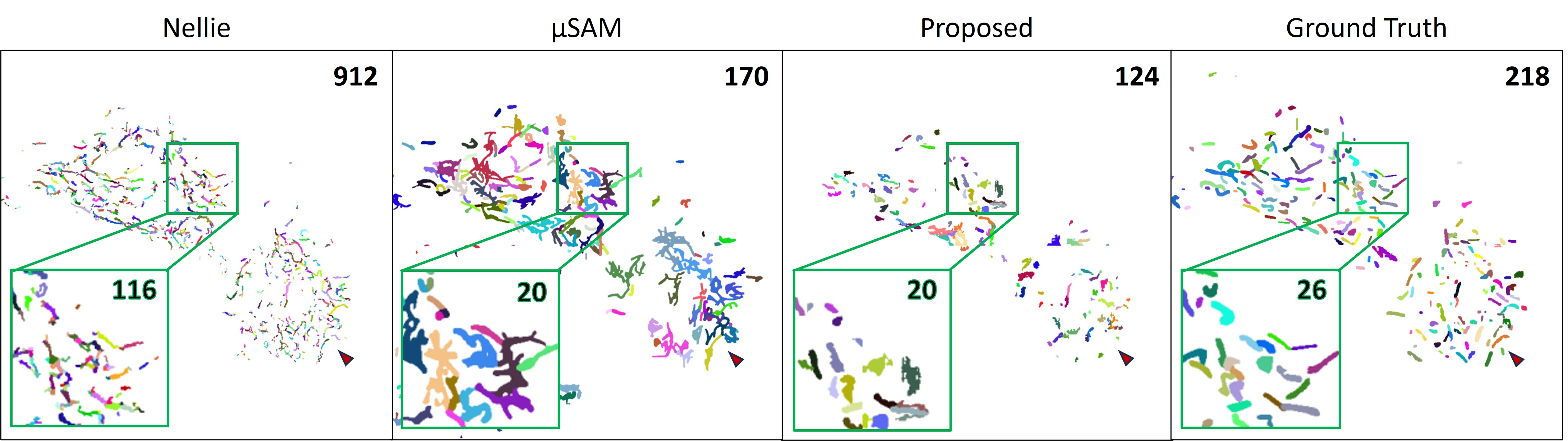}
  \caption{}
\end{subfigure}
\newline
\begin{subfigure}[b]{\textwidth}
  \centering
  \includegraphics[width=\linewidth]{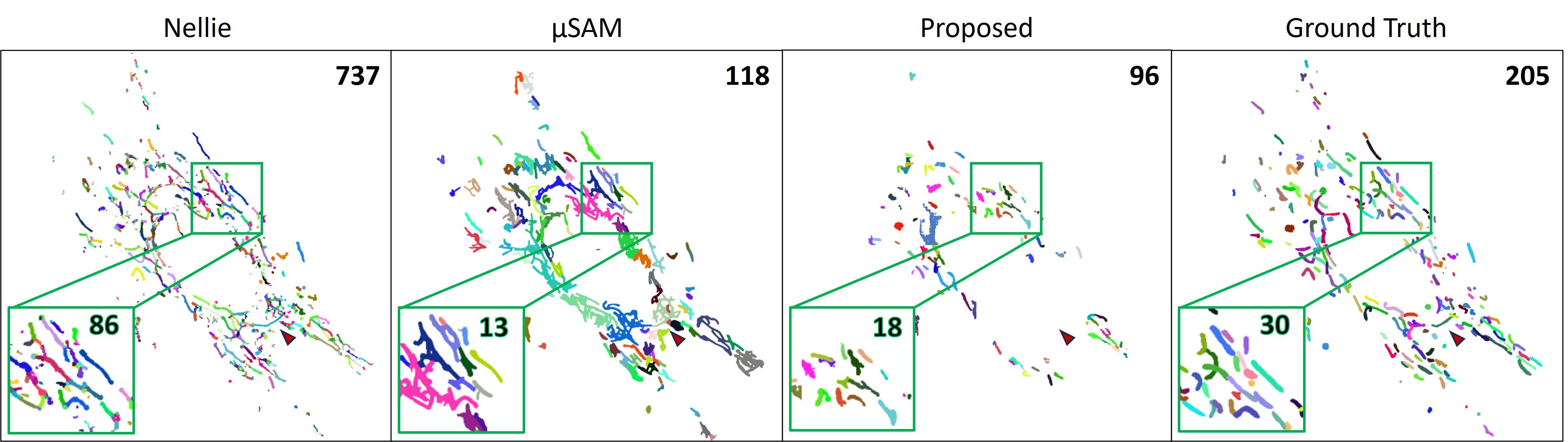}
  \caption{}
\end{subfigure}
\caption{Visual comparison of instance segmentation outputs produced by the different methods on (a) Test image 1, and (b) Test image 2. Numbers inside the zoomed area denote the number of instances detected in the region by each of the methods. Numbers on the top-right corner denote the total count.}
\label{fig:qualitative}
\end{figure*}

Figure \ref{fig:qualitative} shows the segmentation results obtained by each of the methods on the labeled test images, as compared against the manually annotated ground truth. Zoomed-in regions showcase the results obtained by each of the methods on a relatively sharper part of the image. The text inside the boxes denote the instance count detected by each of the methods in the region. 

In terms of output quality, we can see that Nellie's looks the sharpest qualitatively. In both the images, however, we can note that Nellie consistently tends to overestimate the count by a large margin. Whereas the ground truth annotation has counted a total of roughly 200 individual mitochondria instances in both the images, Nellie's count reflects a much higher number. While the output itself adheres well to the underlying structure, Nellie also consistently exhibits a high level of fragmentation, resulting in a large over-estimation of underlying instance count. We can see from the zoomed in region that individual strands of mitochondria are erroneously detected as multiple instances, with junctions and sharp bends as separators. Lastly, Nellie also struggles with noisy regions caused by out-of-focus mitochondria (highlighted with yellow markers in fig. \ref{fig:test_images}). In these regions, we find Nellie erroneously detecting mitochondria instances. Overall, Nellie's overall output matches with the expected foreground the best, but the individual instances are not delineated correctly. This, combined with the high instance count estimate, explains the low precision and high recall scores we saw in section \ref{sec:quant}.
$\mu$SAM and the proposed model, on the other hand, both consistently under-estimate the total instance count, albeit by a much smaller factor as compared to Nellie. In terms of total count, $\mu$SAM is the closest to the ground truth across both the test images. While $\mu$SAM's output covers the entire foreground region very effectively, it also struggles with delineating individual instances. If we look at the zoomed-in region, we can notice that $\mu$SAM consistently tends to cluster up multiple mitochondria in a dense network into one continuous structure. This leads to a heavy under-estimation of the underlying mitochondria count, leading to a low recall. Over-simplification of individual mitochondria in a network also leads to lower precision overall, as it leads to less overlap with individual instances. 

\begin{figure}[htbp]
  \centering
  \includegraphics[width=\columnwidth]{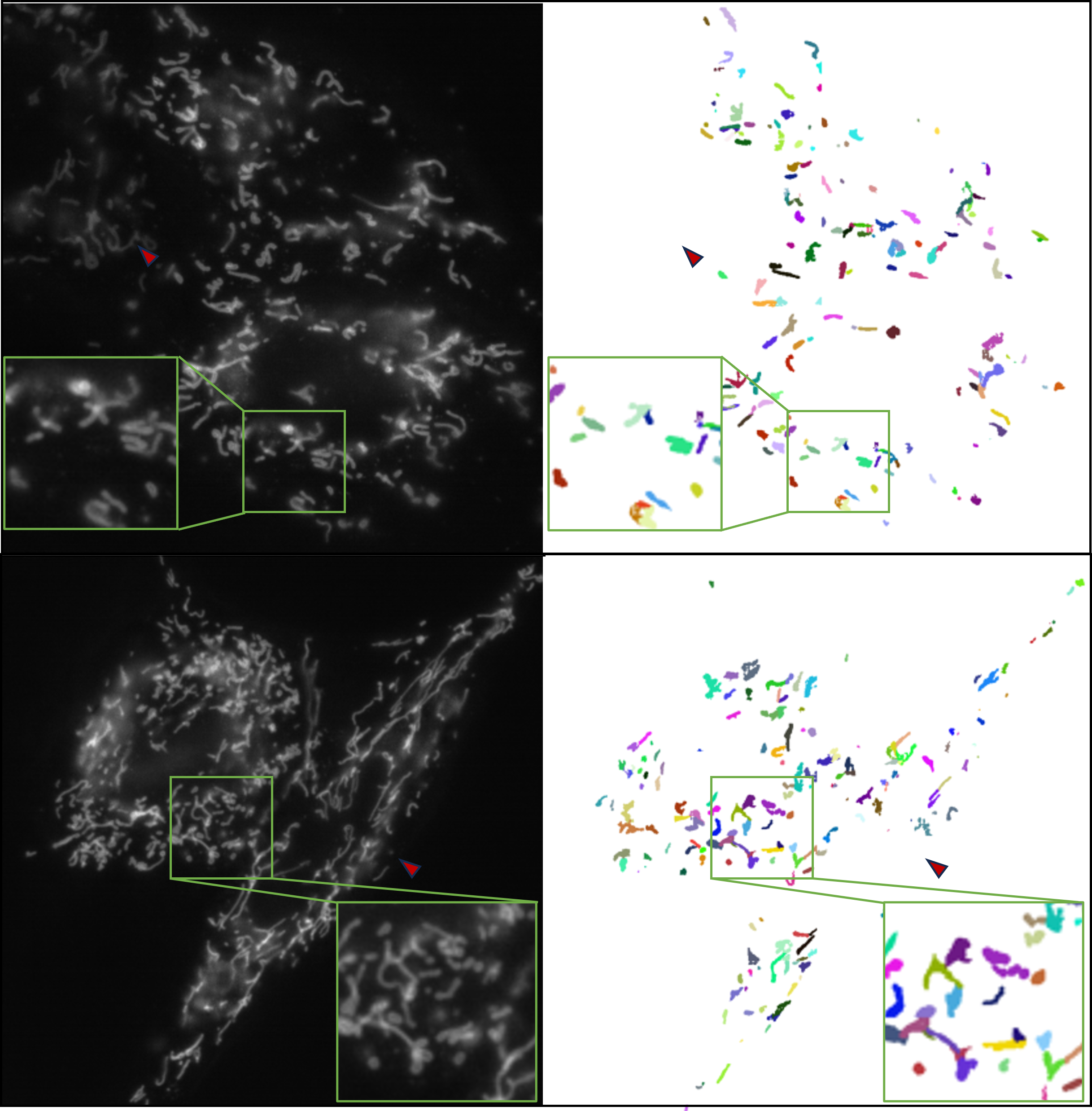}  

\caption{Side-by-side visualization of instance segmentation output produced by the proposed method on unannotated images from \cite{physeg}. Regions marked by green rectangle highlight areas with good detection performance. Red markers highlight areas with missed detections.}
\label{fig:qualitative2}
\end{figure}

Figure \ref{fig:qualitative2} shows the instance segmentation output produced by our model on more unannotated images from \cite{physeg}. From both figures \ref{fig:qualitative} and \ref{fig:qualitative2}, we can note that the proposed model does not cover the entire foreground region the most effectively, as compared to the other two methods. However, the instances that are correctly detected are comparatively more precise. The predicted instance count is closer to the actual count in the zoomed-in regions in fig. \ref{fig:qualitative} and the problems of defragmentation and over-simplification are also both reduced significantly. We can observe this from the zoomed-in region. Individual mitochondria are handled better than compared to Nellie, and defragmentation is greatly prevented. Individual mitochondria are correctly interpreted as being one continuous structure, despite small bends and junctions. On the other hand, in dense networks of mitochondria, over-simplification is mitigated by a good margin and networks get split into individual mitochondria much better than $\mu$SAM. While the proposed model does not cover the entire foreground region as effectively, the delineation of individual instances works the best in comparison to the other two.

Finally, if we look at the low-contrast regions in the input image, like the ones highlighted by red marker in figures \ref{fig:qualitative} and \ref{fig:qualitative2}, we can see that the proposed model is not as effective in detecting mitochondria in these regions. Compared to the other two methods, our proposed model tends to miss detections in these regions.

\subsection{Qualitative Analysis}
\label{sec:qual}

\subsection{Case Study: Morphology-Based Analysis}
Instance segmentation of mitochondria allows for automated analysis of mitochondrial morphology. Such analysis can be useful for studying the results of various biological experiments. In this case study, we analyse a set of fluorescence microscopy images taken under different cell growth conditions. The data for this experiment is taken from \cite{physeg}. The provided dataset contain fluorescence microscopy images of a sample under various different cell growth conditions. Images under "Normal" condition act as the baseline for our analysis. Under "Hypoxia", the overall proportion of dot-type mitochondria is expected to rise from the baseline, accompanied by a relative drop in the proportion of rod-like mitochondria. Finally, under "Hypoxia-ADM", the proportion of dot-type mitochondria is expected to sink close to the baseline again, accompanied with a similar rise in the rod-type mitochondria. 
We use our proposed model to replicate the results obtained in \cite{physeg}. The dataset contains 13 "Normal" images, 12 "Hypoxia" images, and 9 "Hypoxia-ADM" images. For each of these, we first segment all the individual mitochondria using our model. The individual instances are then classified into the dot-type and rod-type based on the area, similar to \cite{physeg}, as shown in eq. \ref{eq:case_study}. One change made here is in the threshold, which is changed from 120 pixels to 200 pixels. We found the higher threshold providing a slightly better class separation with our model. 

\begin{equation}
    {\rm{{Class}_{mitochondria}}}=\left\{\begin{array}{rlrlr}&{\rm{Dot,}}&&\,{{\rm{if}}}\,\,{\rm{area}}\,{\le}\, 200\,{\rm{pixel}}\,&\\ &{\rm{Rod,}}&&\,{{\rm{if}}}\,\,{\rm{area}}\,{>}\,{200}\,{\rm{pixel}}\,\end{array}\right.
\label{eq:case_study}
\end{equation}

\begin{figure}[htbp]
  \centering
  \begin{subfigure}{\linewidth}
    \centering
    \includegraphics[width=\linewidth]{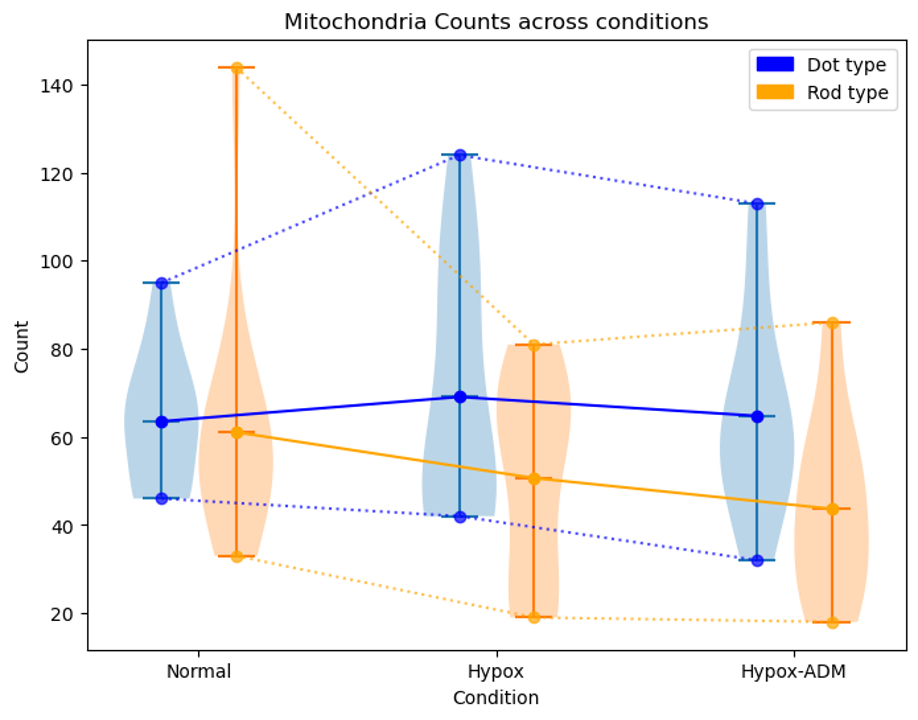}
    \caption{Counts of each type across conditions}
    \label{fig:case_study_a}
  \end{subfigure}
  
  \vspace{0.4cm}
  \begin{subfigure}{\linewidth}
    \centering
    \includegraphics[width=\linewidth]{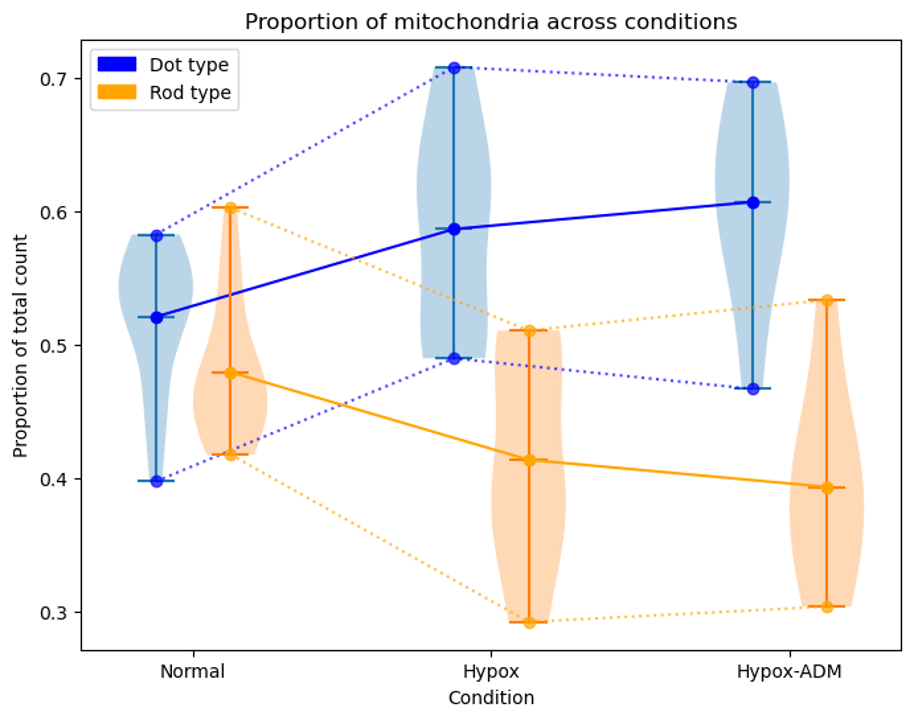}
    \caption{Proportion of total counts of each type across conditions}
    \label{fig:case_study_b}
  \end{subfigure}
  
  \vspace{0.4cm}
  \begin{subfigure}{\linewidth}
    \centering
    \includegraphics[width=\linewidth]{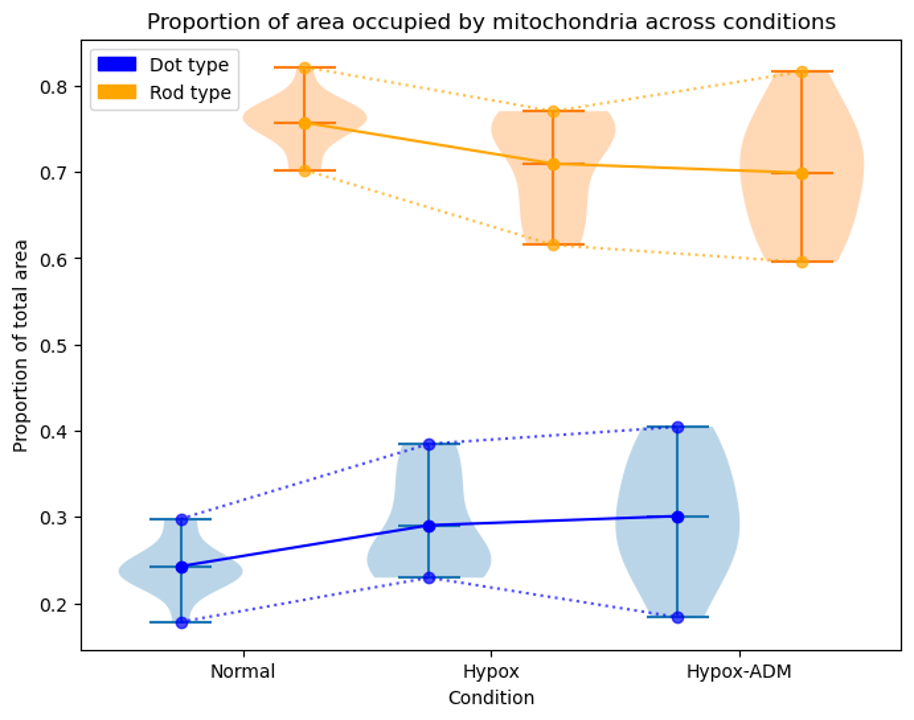}
    \caption{Proportion of areas covered by each type across conditions}
    \label{fig:case_study_c}
  \end{subfigure}
  \caption{Results of morphological analysis across three cell growth conditions - Normal, Hypoxia, and Hypoxia-ADM.}
  \label{fig:case_study}
\end{figure}

Improved instance segmentation capabilities of our approach makes it possible to perform such an analysis. Compared to existing automated techniques that utilise semantic segmentation, having access to instance segmentations provides a deeper insight into individual mitochondrial morphologies in dense networks. In addition to the total counts of each type, we also compute the areas occupied by each type and their proportion in relation to the total foreground area. The counts and areas of each type of mitochondria are accumulated for each cell growth condition, and the statistics are shown in figure \ref{fig:case_study}. 

We conduct three analyses in total. The first one (shown in figure \ref{fig:case_study_a}), is the total counts of each type of mitochondria across conditions. The individual images used in this analysis have different field of views and can contain different amount of mitochondria each. Normalizing per-image statistics takes this variation into account. We show two such analysis in figures \ref{fig:case_study_b} and \ref{fig:case_study_c}, where we plot the \textit{proportions} of total counts and area of each type of mitochondria. Each of these is normalised by the total counts and total area of all instances in the image, respectively. 

Across all the figures, we can note that the produced analysis roughly matches the expected outcome. The expected result for dot-type mitochondria is for each of these statistics to \textit{increase} from normal to hypoxia, and then \textit{decrease} subsequently from hypoxia to hypoxia-ADM. For rod-type mitochondria, the expected result is for each of these statistics to \textit{decrease} from normal to hypoxia, and subsequently \textit{increase} from hypoxia to hypoxia-ADM. From the figure, we can see that rod-type mitochondria (shown in orange) roughly show a V-shaped trend, matching the expected outcome. Dot-type mitochondria, on the other hand, roughly show a $\wedge$-shaped trend, also matching the expected outcome. The trend is the most visible in figure \ref{fig:case_study_c}, where we compare the proportion of total area occupied by each type. The trend becomes less clear with the remaining two analyses, that depend on the counts. As noted in sections \ref{sec:quant} and \ref{sec:qual}, this is consistent with the proposed model's tendency to under-estimate total instance counts. 

%% file: sec/3_finalcopy.tex
\section{Conclusion}
In this paper, we introduced a simulation-supervised training paradigm to adapt the Segment Anything Model (SAM) for robust mitochondria instance segmentation in fluorescence microscopy. By utilizing synthetically generated images for finetuning, we mitigate the ground truth deficiency problem, which traditionally has limited the application of deep learning in the domain. Our evaluations demonstrate that our model outperforms existing state-of-the-art tools like Nellie and $\mu$SAM in precision, mean Dice scores. Furthermore, we validate the practical efficacy of our model through a morphology-based case study on mitophagy, proving its capability to reliably analyse populations of mitochondria under various growth conditions.

While individual instance delineation sees a noticeable improvement, this work is not without its limitations. There is particularly a scope for improvement with regards to its recall scores. This approach does not find \textit{all} the mitochondria in a given image, and some are missed in their entirety. Further progress can be made to increase the total number of detections. The evaluation metric is also arguably limited, and testing on more manually annotated real data could be useful. In conclusion, we hope this work inspires more FM instance segmentation approaches utilizing a simulation-assisted approach similar to ours.